\pdfoutput=1

\documentclass[11pt,a4paper]{article}
\usepackage[hyperref]{emnlp2020}
\usepackage{times}
\usepackage{latexsym}
\usepackage{amssymb}
\usepackage[algo2e, ruled, lined, linesnumbered]{algorithm2e}
\usepackage{multicol,lipsum}
\usepackage{cuted, nccmath}
\usepackage{multirow}
\usepackage{bm}

\usepackage{url}

\aclfinalcopy 

\usepackage{graphicx}
\usepackage{xcolor}

\usepackage{tabularx}
\usepackage{booktabs} 
\usepackage{soul}
\usepackage[T1]{fontenc}
\usepackage{subcaption}
\usepackage{hyperref}
\usepackage{balance}
\usepackage{amsmath}
\usepackage{breakcites}

\usepackage{epstopdf}
\usepackage[latin1]{inputenc}
\usepackage{xstring}

\newif\ifdraft
\drafttrue

\newcommand{\cc}{Common Crawl}
\newcommand{\dataset}{dataset}

\title{CCAligned: A Massive Collection of Cross-Lingual Web-Document Pairs}

\author{Ahmed El-Kishky$^1$ \enskip
Vishrav Chaudhary$^1$ \enskip
Francisco Guzm\'an$^1$ \enskip
Philipp Koehn$^2$ \\
$^1$Facebook AI \enskip $^2$Johns Hopkins University \\
{\tt \{ahelk,vishrav,fguzman\}@fb.com, phi@jhu.edu} \\
}
\date{}
\begin{document}

\maketitle
\begin{abstract}
Cross-lingual document alignment aims to identify pairs of documents in two distinct languages that are of comparable content or translations of each other. In this paper, we exploit the signals embedded in URLs to label web documents at scale with an average precision of 94.5\% across different language pairs. We mine sixty-eight snapshots of the \cc{} corpus and identify web document pairs that are translations of each other. We release a new web \dataset{} consisting of over 392 million URL pairs from \cc{} covering documents in 8144 language pairs of which 137 pairs include English. In addition to curating this massive dataset, we introduce baseline methods that leverage cross-lingual representations to identify aligned documents based on their textual content. Finally, we demonstrate the value of this parallel documents dataset through a downstream task of mining parallel sentences and measuring the quality of machine translations from models trained on this mined data. Our objective in releasing this dataset is to foster new research in cross-lingual NLP across a variety of low, medium, and high-resource languages.
\end{abstract}

\section{Introduction}
Cross-lingual document alignment aims to pair documents such that they are translations or near translations of each other. There are a variety of tasks in natural language processing that consume such parallel cross-lingual data. Traditionally, machine translation approaches have leveraged parallel sentences as training data for use with sequence-to-sequence models. 
Other tasks include cross-lingual information retrieval and cross-lingual document classification. Additionally, cross-lingual data facilitates training cross-lingual representations such as multilingual BERT \cite{devlin2019bert} and XLM \cite{lample2019cross} which are used in many NLP tasks. The availability of high-quality \dataset{}s is necessary to both train and evaluate models across these many tasks.

While it is possible to manually 
label aligned documents across languages, the process is costly and time consuming due to the quadratic search space for document pairs. Additionally, for low-resource languages, identifying these cross-lingual document pairs is more difficult due to their relative scarcity. Furthermore, lack of access to qualified human annotators makes it necessary to have additional quality control in low-resource scenarios \cite{guzman-etal-2019-flores}. 

In this paper, we investigate whether we can rely on weak supervision to generate labels for document pairs. In particular, we focus on the weak signals embedded in the URLs of web documents, that can be used to identify the different translations of a single document across many languages. We preprocess, filter, and apply a set of high-precision hand-crafted rules to automatically sift through a massive collection of 169 billion web documents and identify over a 392 million cross-lingual parallel documents in 8144 language pairs. Of these aligned documents, 292 million are non-English document pairs, and 100 million include English. 
We evaluate the quality of our automatic-annotation setup by running a manual human evaluation on a diverse sample of positively-labeled documents across six language pairs.


 
We also introduce a simple baseline that effectively aligns cross-lingual document pairs using solely textual content and in the presence of detractor documents which may not have any parallel counterpart. 

Finally, we demonstrate the utility of our parallel corpus by demonstrating how these parallel documents can be mined for training data for a downstream machine translation task.

We release the \dataset{} 
consisting of pairs of translated documents represented by URLs extracted from a massive collection web crawls. 
We hope that the size, diversity, and quality of this \dataset{} spurs its use not only as a benchmark for document alignment, but also as a parallel corpus for mining parallel sentences and as supervision for a variety of cross-lingual tasks.

\section{Related Works}

The concept of crawling and mining the web to identify sources of parallel data has been previously explored~\cite{resnik1999mining}. A large body of this work has focused on identifying parallel text from multilingual data obtained from a single source: for example the United Nations General Assembly Resolutions~\cite{rafalovitch2009united,ziemski2016united} or European Parliament parallel corpus~\cite{koehn2005europarl}. 
These parallel corpora were curated from specific, homogeneous sources by examining the content and deriving domain-specific rules for aligning documents. 

Other approaches have identified parallel documents in unstructured web corpora by relying on metadata~\cite{nie1999cross,espla2010combining}. Some of these methods have focused on publication date and other temporal heuristics to aid in identifying parallel documents~\cite{munteanu2005improving,munteanu2006extracting,udupa2009mint,do2009mining,abdui2009use}. However, temporal features can be sparse, noisy, and unreliable. A different class of alignment methods rely on document structure~\cite{resnik2003web,chen2000parallel}. 

In the WMT-2016 bilingual document alignment shared task~\cite{buck2016findings}, many techniques applied retrieval and matching on translated 5-grams ~\cite{dara2016yoda} to query, retrieve, and align documents. Similar methods for generating candidates by retrieving matches based on the least frequent bi-lingual 5-grams have been proposed~\cite{gomes2016first} with the insight that rare snippets are more informative. Both of these candidates rely on high-quality translation systems to translate either the source or the target. Such models may not exist, especially for low-resource language directions. The application of alignment to a variety of languages was not explored in WMT-2016 which only considered English to French document alignment -- a high-resource direction.

Recently, the use of neural embedding methods has been explored for bilingual alignment of text at the sentence and document level. \citet{guo-EtAl:2019:WMT2} propose using hierarchical document embeddings, constructed from sentence embeddings, for bilingual document alignment.

\section{Dataset Creation and Description}

The \cc{} corpus is a publicly available crawl of the web. With a new snapshot uploaded each month, and over 2 billion pages released in each snapshot, this data is a vast resource with content across a large number of domains and languages.  Previous works have leveraged the data from \cc{} for mining ngram counts to perform language modeling~\cite{Buck-commoncrawl}. Other works~\cite{smith2013dirt} have mined \cc{} for bitexts for machine translation. However, this mining was performed on a small scale. For our \dataset, we use 68 snapshots published from 2013 to 2020 which is vastly larger than previous works.


\subsection{Preprocessing}
The first step in preprocessing the data is deduplication. While investigating combining many \cc{} snapshots, we found duplicate URLs both within an individual snapshot and almost always across snapshots. As our data curation method relies on unique URLs for each web document, we apply a heuristic to ensure each URL appears once within the final cleaned data. The first step is to normalize each URL; we perform this by simply removing the protocol and host name (e.g., {\tt https://www.aaa.com} $\rightarrow$ {\tt aaa.com}). Upon normalization, for each URL that appears more than once, we select the instance that possesses the longest document content. This heuristic assumes that occasionally, content is (1) deleted and gets shorter or (2) is amended and gets longer. In this case, it is preferable to operate on the larger content. Starting from 68 \cc{} snapshots with a raw document count of 169.4 billion documents, upon deduplication, the resultant corpus is approximately 29.6 billion web documents from 107.8 million distinct web domains -- a $83\%$ reduction from the raw corpus.

\subsection{Language ID \& URL Matching}
The next step in the pipeline is to tag each document with the dominant language identifier. We utilize a lightweight text classifier such as FastText~\cite{joulin2017bag} that has been trained to detect a variety of languages. 

To identify pairs of cross-lingual documents, we apply a high-precision, low recall heuristic to assess whether two URLs represent web pages that are translations of each other. This heuristic presumes that two URLs, with high probability, refer to pages that are translations of each other if both can be transformed into the same string after stripping language identifiers. To improve recall, we allow matches where only one of the pair of URLs contain a language identifier e.g., \url{https://statmt.org} would be a match to \url{https://fr-fr.statmt.org}. We further ensure that these matches are high-precision by verifying that the language identifier stripped from the URL reflects the language of the web document as predicted by the language identifier; otherwise it is discarded.

\begin{table}[!ht]
    \small
\begin{center}
\begin{tabular}{ l  l }

      \toprule
      \textbf{Source URL} & \textbf{Target URL}\\
      \midrule
       \small\textbf{eng.}\nolinkurl{aaa.com} & \small\url{aaa.com}\\
      \small\nolinkurl{aaa.com/}\textbf{en-gb}\nolinkurl{/b}	 & \small\nolinkurl{aaa.com/}\textbf{zh-cn}\nolinkurl{/b}\\
     \small\nolinkurl{aaa.com/}\textbf{English}\nolinkurl{/b}	 & \small\nolinkurl{aaa.com/}\textbf{Yoruba}\nolinkurl{/b}\\
     \small\nolinkurl{aaa.com/b/}\textbf{en}& \small\nolinkurl{aaa.com/b/}\textbf{vi}\\
     \small\nolinkurl{aaa.com/b/}& \small\textbf{thai.}\nolinkurl{aaa.com/b/}\\
     \small\nolinkurl{aaa.com/b}\textbf{\&lang=english}& \small\nolinkurl{aaa.com/b}\textbf{\&lang=arabic}\\
     \small\nolinkurl{aaa.com/b}\textbf{?lang=en}& \small\nolinkurl{aaa.com/b}\textbf{?lang=fr}\\
     \small\nolinkurl{aaa.com/b}& \small\nolinkurl{aaa.com/b}\textbf{?lang=1}\\
      \bottomrule

\end{tabular}
\caption{URL matching via language identifiers.\label{tab:striplang}}
 \end{center}
\end{table}

Table~\ref{tab:striplang} shows a few examples of pairs of aligned URLs. Alignment is performed by normalizing  each URL by stripping its present language identifiers. Extra care is taken to ensure relevant indicators such as \textit{/}, \textit{\&}, and \textit{?} are stripped as well to ensure proper alignment between URLs. 
For reproducibility, we publish an explicit list of patterns used along with the code implementing the pattern matching alongside the dataset. 

Given these rules and restrictions, we mined over 392 million aligned documents (100M with English and 292M without English) across 68 \cc{} snapshots. We assess the efficacy of this rule-based alignment in the next section.

We select a small subset of the original 68 Common Crawl snapshots to use for evaluating baseline document alignment methods. These 121K documents contain English and non-English documents from 450 web domains. Of these documents, 17.5K pairs are aligned as defined in our URL-aligned dataset. We release this test set as a tractable collection of documents on which to benchmark different alignment methods.


\section{Dataset Evaluation}
\label{sec:quality}
In this section, we analyze the quality of our cross-lingual URL-aligned \dataset{}. We assesses the quality by measuring the precision of a representative sample of the URL-aligned data to human-annotated alignment judgments. 

\subsection{Dataset Quality Evaluation}
To assess the quality of the cross-lingual document pairs obtained by our method, 
we recruit human annotators to evaluate the alignments and provide an assessment of whether the documents in the pair are total or partial translations of each other. To perform the evaluation, we first selected six languages from various language families, scripts, and levels of resource availability. For each language, we randomly identified 30 pairs of URLs from different web domains aligned into English for a total of 180 pairs. To gather pairs from a diverse set of websites, each URL pair is selected from a distinct web domain.

Then, we tasked twelve human annotators (bilingual or trilingual) to annotate URL pairs. These annotators are fluent in the pairs of languages being assessed and rate the alignment by loading the two web-pages corresponding to each URL in a pair side by side and assessing whether or not the content rendered is both comparable \textit{and} in the correctly tagged language. Each URL pair was evaluated by three human annotators to add a level of redundancy and measure annotator agreement. Note that the evaluation was performed in early November 2019, therefore for some of the pairs in the set, the document content had changed from the time when the \cc{} snapshot was generated. 

\begin{table}[!ht]
\small
    \centering
    \begin{tabular}{l  l  r r r }
        \toprule
        & \textbf{Language} & \textbf{P$_{maj}$} &  \textbf{K}$\bm{\alpha}$ & \textbf{P$_{adj}$}  \\\midrule
       \multirow{2}{*}{{\bf High} } &  German  & 90.0 & 0.74 & 96.7  \\
       & Chinese    & 86.7 &0.68  &  93.3 
       \\\midrule
       \multirow{2}{*}{{\bf Mid}}  & Arabic  & 83.3 & 0.72 &  90.0 \\
        & Romanian   & 76.7 &0.50  & 96.7 \\\midrule
        \multirow{2}{*}{{\bf Low }} &Estonian  & 83.3 &0.68  & 90.0  \\
        & Burmese   & 86.7 &0.88  &   100.0 \\
        \midrule
        & \textbf{Avg} & 84.4 & 0.70 & 94.5
        \\\bottomrule
        
    \end{tabular}
    \caption{Human evaluation of documents of different languages aligned to English. Languages are classified as high, medium or low resource based on the amount of mined documents. We report the majority-vote precision \textbf{P}$_{maj}$ and the precision after accounting for experimental error \textbf{P}$_{adj}$. Additionally, we report \textbf{K}$\bm{\alpha}$, a measure of inter-rater agreement.}
    \label{tab:quality_human}
\end{table}

In Table~\ref{tab:quality_human}, we report the precision of our method to generate URL-aligned documents. As individual raters may have differing opinions on what constitutes a cross-lingual comparable document, we report results according to the majority vote. In addition, we report the inter-rating agreement among annotators as measured by the Krippendorff Alpha~\cite{krippendorff2011computing} of the annotations. After observing annotator comments and performing a round of error analysis on the pairs identified as misaligned, we identified the following reasons: (1) In $40\%$ of the cases, the content of the rendered web-page has changed since the \cc{} snapshot was generated or the URL redirects the user to a different page, while the original \cc{} is a parallel document; (2) In $20\%$ of cases, the content in one of the parallel documents appears to be much shorter than the document in the original (dominant) language but the message is the same, which many annotators didn't consider the document pairs as translations of each other; (3) In $10\%$ of cases the majority of dynamic content within a document pair appears to be in the same language and only boilerplate text such as columns and title are translations; and the remaining $30\%$ are truly non-comparable documents due to a myriad of different reasons.  
To alleviate the issues introduced by (1) due to experimental setup (i.e. using a freshly rendered web-page) and (2) due to guidelines issues (i.e. partial translations), we sent those cases for an additional round of annotation. The resulting \emph{adjusted} precision after the second round is observed as $P_{adj}$.

Overall, we observe that the URL pairs in \cc{} appear to adhere to human-standards of comparability with a majority of measured directions achieving precision of over 90\%.
\section{Document Alignment Experiments}
In Section~\ref{sec:quality}, we verify the quality of the URL-aligned \dataset{} through human-evaluation. In this section, we treat the URL-aligned \dataset{} as a high-precision, low-recall \dataset{} and evaluate baselines that score document pairs based on content rather than URL information. The scored document pairs are then aligned via a greedy bipartite matching algorithm. The resulting alignments are evaluated on a subset of the URL-aligned \dataset{} which is treated as ground truth.

\subsection{Problem Definition}

Given a set of source documents, $D_s$ and a set of target documents $D_t$, cross-lingual document alignment aims to find the largest set of pairs of documents from source to target $(d_s, d_t)$ where $d_s \in D_s$ and $d_t \in D_t$ such that each source document and target document can only be used in at most a single pair.

To find the best possible mapping between $D_s$ and $D_t$ we require two components: 1) a similarity function $\phi(d_s, d_t)$ which is used to score a set of candidate documents according to their relatedness; and 2) an alignment or matching algorithm which uses the scores for each of the pairs in $D_s \times D_t$ to produce an alignment of size $min(|D_s|, |D_t|)$ representing the best mapping according to $\phi(d_s, d_t)$.




In the remainder of this section, we introduce our proposed baseline  document pair similarity functions and a simple matching algorithm that 
aligns source and target documents. 




\subsection{Embedding-Based Document Similarity}
To guide the alignment algorithm, a notion of cross-lingual document similarity is necessary. This score should capture the fact that two documents are \textit{semantically} similar despite having some or all of their content in different languages. We describe three simple language-agnostic document embedding methods. These embeddings leverage LASER\footnote{\url{https://github.com/facebookresearch/LASER}}~\cite{artetxe2019massively}, a multilingual sentence representation that uses byte-pair encoding to share the same vocabulary among all languages and trained on parallel sentences pulled from Europarl, United Nations, OpenSubtitles2018, Global Voices, Tanzil and Tatoeba
corpus, covering 93 languages.

\paragraph{Direct Embedding}
The first baseline, Direct Embedding (DE) uses a standard cross-lingual encoder to directly embed each document. Each document $d$ has its dense vector representation $\mathbf{v}_{d}$ computed by applying the open-source cross-lingual LASER encoder to its \textit{full textual content}. 

\paragraph{Sentence Average Embedding}
The second baseline, sentence average (SA), performs document embedding by first decomposing each document into sentences, embedding each sentence, then combining these sentence representations. Given a document $d$, we segment it into a list of sentences $\{s_{i}\}_{i=1}^{n}$. This time, the LASER encoder is used to encode each sentence $s_{i}$ into a dense vector $\mathbf{v}_{s_{i}}$. After embedding each sentence in a document, document embedding is performed by averaging the sentence vectors into a document vector $\mathbf{v}_{d}$ as follows:

\begin{align}
    \mathbf{v}_{d} = \frac{1}{n}\sum\limits_{i=1}^{n}\mathbf{v}_{s_{i}}
\end{align}

\noindent
\paragraph{Weighted Average}
For the final baseline, we extend the simple sentence averaging to incorporate importance weights for each sentence. To compute this weighted average (WA), we once again embed sentences, however, each sentence's embedding vector is scaled by an importance weight before being averaged to construct the document vector $\mathbf{v}_{d}$ as follows:

\begin{align}
    \mathbf{v}_{d} = \frac{1}{n}\sum\limits_{i=1}^{n}\mathbf{w}_{s_{i}} \times \mathbf{v}_{s_{i}}
\end{align}

We investigate three potential weighting schemes that draw inspiration from tf-idf~\cite{ramos2003using}. This weighting scheme is reminiscent of the use of tf-idf to determine word relevance, but instead sentence length and inverse document frequency of a sentence within a web-domain is used. In our experiments we compute $w_i$ for sentence $s_i$ in document $d$ in these three ways.

For the first weighting scheme, we posit that longer sentences should be assigned larger weighting than shorter sentences. To capture this, we weight each sentence by the number of tokens in the sentence relative to the total number of tokens in the entire document. We compute this sentence length (SL) weighting scheme as follows:
\begin{equation}
SL_{s_i} = \frac{|s_i|}{\underset{s\in d}{\sum} count(s) \times |s|}
\end{equation}
Note that this sentence-length weighting is analogous to term-frequency in tf-idf, which is used to assess the importance of a term within a document.

The second insight we investigate is that sentences and text segments that are more frequent in a web-domain are likely boilerplate or less informative segments and should be down-weighted. As such we compute inverse document frequency for each sentence as an alternative weighting scheme. For specific webdomain $D$ we compute IDF as follows:
\begin{equation}
IDF_{s_i} =  \log \frac{N + 1}{1 + |\{d \in D: s \in d\}|}
\end{equation}
where $N$ is the total number of web-documents in the web domain $D$, and $|\{d \in D: s \in d\}|$ is the number of documents where the sentence $s$ occurs.

Finally, similarly to combining term-frequency and inverse document frequency in tf-idf, we combine the two sentence-weighting schemes into a third that captures both insights as follows:
\begin{equation}
SLIDF_{s_i} = SL_{s_i} \times IDF_{s_i}
\end{equation}

\noindent
\paragraph{Scoring}
Using the dense document representations for each document from the source and target sets, the next step is to score pairs to evaluate how semantically similar documents are. Given two documents $a$ and $b$, we compute their semantic similarity using a cosine similarity score: 
\begin{equation}
sim(a,b) = \frac{\mathbf{v}_{a} \cdot \mathbf{v}_{b}}{||\mathbf{v}_{a}||~ ||\mathbf{v}_{b}||}
\end{equation}

\subsection{Competitive Matching Alignment}
Using the baseline scoring function, we score all document pairs in the same web domain that belong to the source and target languages respectively. As such, for any given domain, each document in the source document set, $D_s$ is paired with each document in the target set, $D_t$, yielding $D_s \times D_t$ scored pairs -- a fully connected bipartite graph. Just like in~\cite{buck2016quick}, the expected output assumes that each page in the non-dominant language has a translated or comparable counterpart. This yields a $min(|D_s|, |D_t|)$ expected number of aligned pairs.

While an optimal matching maximizing scoring can be solved using the Hungarian algorithm~\cite{munkres1957algorithms}, the complexity of this algorithm is $\mathcal{O}(max(|D_s||D_t|)^3)$ which is intractable to even moderately sized web domains. As such, similar to the work in~~\cite{buck2016quick}, a one-to-one matching between English and non-English documents is enforced by applying, competitive matching, a greedy bipartite matching algorithm. 

\begin{algorithm2e}[ht]
\caption{Competitive Matching}
\label{alg:greedy}
\Indm
\small
       \KwIn{$P = \{(d_s, d_t) | d_s \in D_s, d_t \in D_t\}$} 
       \KwOut{$P'= \{(d_{s,i},d_{t,i}), . . .\} \subset P$}
\Indp
       \BlankLine
    $scored$ $\gets \{(p, score(p)) \text{ for } p \in P\}$ \\
	$sorted$ $\gets sort(scored) \text{ in descending order } $\\
	aligned $\gets \varnothing$\\
	$S_s \gets \varnothing$\\
	$S_t \gets \varnothing$ \\
	\For{$d_s, d_t \in$ sorted}{
		if $d_s \notin S_s \land d_t \notin S_t$ 
		{
		   $aligned \gets aligned \cup \{(d_s, d_t)\}$ \\
		   $S_s \gets S_s \cup d_s$ \\
		   $S_t \gets S_t \cup d_t$ \\
		}	
	}	
	  \textbf{return} aligned
\end{algorithm2e}
\normalsize

In Algorithm~\ref{alg:greedy}, each candidate document pair is scored using the document similarity scoring function. These candidates are then sorted in order of most similar to least similar using their numerical score. The algorithm then iteratively chooses a document pair with the highest score as long as the $d_s$ and $d_t$ of each pair have not been used in a previous (higher scoring) pair. The algorithm terminates when $min(|D_s|, |D_t|)$ pairs have been selected. Unlike the Hungarian algorithm, the runtime complexity is a more tractable $\mathcal{O}(|D_s||D_t|\times\log(|D_s||D_t|))$ which is dominated by the cost of sorting all candidate pairs.

\subsection{Baseline Results}
We evaluate the baseline scoring by aligning the documents from a subset of the 68 \cc{} snapshots. We score document pairs within the same webdomain using the DE, SA, and WA embedding methods respectively and computing cosine similarity between their representations. For the alignment, we report the performance for each embedding method after applying our competitive matching alignment algorithm as described in Algorithm~\ref{alg:greedy}.

Recall (i.e. what percentage of the aligned pages in the test set are found) is computed on a test-set consisting of pairs from the URL-aligned documents, which we verified have high-precision and we treat as the ground-truth test set.

\begin{table*}[t]
\begin{minipage}[b]{0.28\linewidth}\centering
    \scriptsize
    \subcaptionbox{High-resource directions. \label{subtable:highresource_baseline}}{
    \centering
    \setlength\tabcolsep{3.0pt}
        \begin{tabular}{l  c  c  c  c c}
            \toprule
            & \multicolumn{5}{c}{{\bf Recall}}\\\cmidrule{2-6}
            \textbf{Language} & \textbf{DE} & \textbf{SA} & \textbf{SL} & \textbf{IDF} & \textbf{SLIDF} \\\midrule
            French &  0.39 & \textbf{0.84}  & 0.83 & 0.82 & \textbf{0.84}\\
            Spanish &  0.34 & 0.53 & 0.55 & \textbf{0.58} & 0.57\\
            Russian  &  0.06 & 0.48 & 0.50 & \textbf{0.61} & 0.60\\
            German &  0.52 & 0.74 & \textbf{0.76} & 0.74 & \textbf{0.76}\\
            Italian &  0.22 & 0.54 & 0.55 & 0.55 & \textbf{0.57}\\
            Portuguese  &  0.17 & 0.36 & 0.39 & 0.33 & \textbf{0.40}\\
            Dutch  &  0.28 & 0.51 & 0.54 & 0.52 & \textbf{0.56}\\
            Indonesian  & 0.11 & 0.36 & \textbf{0.48} & 0.43 & \textbf{0.48}\\
            Polish  & 0.17 & 0.38 & 0.41 & \textbf{0.44} & 0.42\\
            Turkish  &  0.12 & 0.30 & 0.34 & \textbf{0.45} & 0.41\\
            Swedish  & 0.19 & 0.37 & 0.37 & 0.38 & \textbf{0.39}\\
            Danish  &  0.27 & 0.46 & 0.65 & 0.60 & \textbf{0.67}\\
            Czech  &  0.15 & 0.36 & \textbf{0.41} & 0.32 & \textbf{0.41} \\
            Bulgarian  &  0.07 & 0.34 & 0.37 & 0.40 & \textbf{0.44}\\
            Finnish  &  0.06 & 0.24 & 0.32 & 0.43 & \textbf{0.44}\\
            Norwegian  & 0.13 & 0.26 & 0.33 & 0.33 & \textbf{0.38}\\
            \midrule 
            \textbf{Macro-AVG} &  0.20 & 0.41 & 0.45 & 0.47 & \textbf{0.49}\\
            \bottomrule 
        \end{tabular}
    }
    \end{minipage}
    \hspace{0.5cm}
    \begin{minipage}[b]{0.28\linewidth}\centering
    \scriptsize
        \subcaptionbox{Mid-resource directions. \label{subtable:mindresource_baseline}}{
        \setlength\tabcolsep{3.0pt} 
        \begin{tabular}{l  c  c  c  c c}
            \toprule
            & \multicolumn{5}{c}{{\bf Recall}}\\\cmidrule{2-6}
            \textbf{Language} & \textbf{DE} & \textbf{SA} & \textbf{SL} & \textbf{IDF} & \textbf{SLIDF}\\\midrule
            Romanian  &  0.15 & 0.39 & 0.40 & 0.40 & \textbf{0.41}\\
            Vietnamese  &  0.06 & 0.13 & 0.18 & 0.15 & \textbf{0.23}\\
            Ukrainian  &  0.05 & 0.49 & 0.70 & 0.70 & \textbf{0.74}\\
            Greek  &  0.05 & 0.22 & 0.24 & \textbf{0.34} & 0.30\\
            Korean  &  0.06 & 0.49 & 0.47 & 0.49 & \textbf{0.51}\\
            Arabic  &  0.04 & 0.26 & 0.46 & 0.42 & \textbf{0.51}\\
            Croatian  &  0.16 & 0.32 & 0.36 & 0.34 & \textbf{0.36}\\
            Slovak  &  0.20 & 0.37  & \textbf{0.44} & 0.41 & 0.42\\
            Thai  &  0.02 & 0.15 & 0.28 & 0.19 & \textbf{0.35}\\
            Hebrew  &  0.05 & 0.19 & 0.30 & 0.27 & \textbf{0.33}\\
            Hindi  &  0.04 & 0.03 & 0.33 & 0.28 & \textbf{0.43}\\
            Hungarian  & 0.15 & 0.41 & 0.39 & 0.39 & \textbf{0.46}\\
            Lithuanian  & 0.11 & 0.61 & 0.72 & 0.74 & \textbf{0.80}\\

            Slovenian  &  0.13 & 0.20 & 0.26 & 0.31 & \textbf{0.33}\\
            Farsi  &  0.06 & 0.22 & 0.37 & 0.40 & \textbf{0.49}\\
            \\
            
           \midrule 
           \textbf{Macro-AVG} &  0.09 & 0.28 & 0.39 & 0.39 & \textbf{0.44}\\
            \bottomrule 
        \end{tabular}
    }
    \end{minipage}
        \hspace{0.5cm}
    \begin{minipage}[b]{0.28\linewidth}\centering
    \scriptsize
    \subcaptionbox{Low-resource directions. \label{subtable:lowresource_baseline}}{
        \setlength\tabcolsep{3.0pt}
        \begin{tabular}{l  c  c  c  c c}
            \toprule
            & \multicolumn{5}{c}{{\bf Recall}}\\\cmidrule{2-6}
            \textbf{Language} & \textbf{DE} & \textbf{SA} & \textbf{SL} & \textbf{IDF} & \textbf{SLIDF} \\\midrule
            
            Estonian  &  0.28 & 0.57 & 0.62 & 0.58  & \textbf{0.64}\\

            Bengali & 0.05 & 0.47 & \textbf{0.59} & 0.51  & 0.58\\
            Albanian  &  0.23 & 0.56 & 0.60 & 0.57  & \textbf{0.61}\\
            Macedonian  &  0.02 & 0.16 & \textbf{0.22} & 0.19  & 0.08\\
            Urdu  &  0.06 & \textbf{0.29} & 0.23 & 0.27  & 0.24\\
            Serbian  &  0.06 & 0.46 & \textbf{0.58} & 0.47  & 0.56\\
            Azerbaijani	   & 0.08 & 0.27 & 0.28 & \textbf{0.34}  & 0.27\\
            Armenian  &  0.02 & 0.08 & 0.13 & 0.12  & \textbf{0.17}\\
            Belarusian  &  0.07 & 0.26 & 0.44 & 0.36  & \textbf{0.51}\\
            Georgian  &  0.06 & 0.18 & 0.23 & \textbf{0.25}  & \textbf{0.25}\\
            Tamil  &  0.02 & 0.13 & 0.19 & 0.23  & \textbf{0.34}\\
            Marathi  &  0.02 & 0.13 & \textbf{0.20} & 0.10  & 0.16\\
            Kazakh  &  0.05 & 0.16 & 0.24 & 0.25  & \textbf{0.33}\\
            Mongolian  &  0.03 & 0.01 & 0.05 & 0.10  & \textbf{0.22}\\

            Burmese  & 0.01 & \textbf{0.35} & 0.18 & 0.08  & 0.26\\
            Bosnian  & 0.18 & 0.49 & 0.64 & 0.50  & \textbf{0.65}\\
            
            \midrule 
            \textbf{Macro-AVG} & 0.08 & 0.29 & 0.34 & 0.31  & \textbf{0.37}\\
            \bottomrule 
        \end{tabular}
        }

    \end{minipage}
    \caption{Recall from \cc{} documents aligned using the baseline content-based alignment methods.} 
    \label{tab:results_baselines}
\end{table*}
\normalsize

We show the alignment results in Table~\ref{tab:results_baselines}. Comparing DE which directly applies LASER to the entirety of the document content, we see that performance is significantly worse than SA. We suspect this may be the case for two reasons (1) the LSTM-based sentence encoders may suffer at representing the semantic meaning of long documents as the model is originally trained on sentences (2) there may be noisy boiler plate content at the beginning of each web document that is less useful semantically but dominates the representation. Averaging improves the document representation by giving each section of the document equivalent contribution to the final document representation. 

When we investigate methods to assign importance weighting to different portions of the document when constructing the document representation, we confirm that importance-weighting segments improves document alignment recall. As seen in Table~\ref{tab:results_baselines} for SL, weighting longer sentences proportionally more than shorter sentences almost universally outperforms SA; on average weighting by SL improves alignment recall by 17\%, 39\%, and 10\% over SA for low, mid, and high-resource pairs. Similarly down-weighting sentences that occur frequently in the web-domain improves performance by 7\%, 39\%, and 15\% over SA. However combining both weighting schemes (SLIDF) outperforms both yielding a 28\%, 57\%, and 20\% improvement. 

Additionally, we observe that as the resource availability increases, the baselines perform better. This may be due to the fact that the LASER embedding models were trained with parallel data and more high-resource parallel sentences were used for training. Finally, it appears that across low, mid, and high-resource directions, European languages appear to be consistently easier to align than non-European languages. For example, Albanian, Serbian, Bosnian, and Belarusian were all aligned with over 0.50 recall despite being low-resource. This may be a by-product of the shared semantic subword vocabulary used by LASER improving performance for low-resource European languages due to their linguistic similarity with the many high-resource European languages.

\section{Case Study: MT as an Application}
In the Section \ref{sec:quality} we did a manual evaluation to gauge the quality of our high-precision alignment method. However, the study is small and does not shed light on the quality of the full dataset.  

Therefore, to assess the quality of the aligned document corpus, we propose a downstream task that leverages the aligned document data as a source of supervision for a massively multilingual machine translation task.
Our intuition is as follows: cross-lingual document pairs can be used to extract translations that in turn can be used for downstream training of sequence-to-sequence translation systems. Given a set of parallel documents, our expectation is that any reasonable mining algorithm would be able to extract high-quality translations, while the opposite is non-parallel documents. 
While this is not a standard MT experiment, using MT for downstream evaluation has previously been used to evaluate the quality of sentence filtering and mining approaches. 
For instance, in the WMT Parallel Corpus Filtering tasks \cite{koehn-etal-2018-findings, koehn-etal-2019-findings} the downstream performance on a translation task is used as a proxy to determine the quality of a similarity (or filtering) function.

Here, we use MT to evaluate the quality of the mined sentences, and compare it to other independently mined corpora.
Our expectation is that, if successful, our approach should be able to mine parallel sentences that are of comparable quality to recent approaches that leverage Wikipedia \cite{schwenk2019wikimatrix} and ParaCrawl\footnote{\url{https://paracrawl.eu/}}~\cite{espla2019paracrawl}, reliable sources of comparable documents.

\paragraph{Sentence mining} The first step is to decompose and mine the aligned document corpus for parallel sentences. For simplicity, we segment each document solely on new lines. Given each document pair's decomposition into sentences, we seek to align sentences within each pair of documents. We then aggregate the parallel sentences across all document pairs to form a parallel sentences \dataset{} suitable for training machine translation models.

We use the open-source LASER toolkit~\cite{schwenk2018filtering} with  the margin-based filtering criterion to mine sentences, as this method has been shown to accurately align sentences for across a variety of low, mid, and high-resource directions~\cite{schwenk2019wikimatrix,chaudhary-EtAl:2019:WMT}.  We use the extracted bitexts for training our NMT systems. 

\begin{table*}[ht]

    \footnotesize
        \begin{tabular}{l  c  c  c  c  c c c c c c c c c}
            \toprule
             & \multicolumn{2}{c}{{\bf Danish}} & \multicolumn{2}{c}{{\bf Croatian}} & \multicolumn{2}{c}{{\bf Slovenian}} & \multicolumn{2}{c}{{\bf Slovak}} & \multicolumn{2}{c}{{\bf Lithuanian}} & \multicolumn{2}{c}{{\bf Estonian}} \\\cmidrule{2-13}
            \textbf{Language} & \textbf{En--\emph{x}} & \textbf{\emph{x}--En} & \textbf{En--\emph{x}} & \textbf{\emph{x}--En} & \textbf{En--\emph{x}} & \textbf{\emph{x}--En} & \textbf{En--\emph{x}} & \textbf{\emph{x}--En} & \textbf{En--\emph{x}} & \textbf{\emph{x}--En} & \textbf{En--\emph{x}} & \textbf{\emph{x}--En}\\\midrule
            WikiMatrix   & 30.9 & 32.9  & 18.8 & 22.4 & 16.5 & 17.3 & 13.8 & 16.9 & - & - & - & -\\
            ParaCrawl    & \textbf{37.3} & \textbf{39.8}  & 23.0 & 29.0 & \textbf{20.4} & \textbf{22.7} & \textbf{20.4} & \textbf{24.3} & 16.5 & \textbf{22.5} & \textbf{15.6} & 19.4\\
            CCAligned    & 37.1 & 38.2  & \textbf{23.5} & \textbf{29.3} & 19.6 & 21.7 & \textbf{20.4} & 24.2 & \textbf{16.7} & 21.8 & \textbf{15.6} & \textbf{20.0}\\
            \bottomrule 
        \end{tabular}

    \caption{BLEU scores of NMT models trained on bitext data mined from various web-sources including Wikipedia, ParaCrawl, and our CCAligned document set evaluated on TED Talk test sets.}
    \label{tab:results_mt}
\end{table*}
\normalsize
 \paragraph{Experimental setup } 
 First the data is processed to induce a 5000 subword vocabulary using SentencePiece~\cite{kudo2018sentencepiece}. The model used is a transformer model from fairseq~\cite{ott2019fairseq} with embeddings shared in the encoder and decoder, 5 encoder and decoder layers with dimensionality 512 are used, encoder and decoder FFN with 2 attention heads each with an embedding dimension of 2048 are used along with encoder and decoder normalization. Dropout of 0.4, attention dropout of 0.2 and relu dropout of 0.2 are applied. The adam optimizer is used to train the model for up to 20 epochs by optimizing a smoothed-cross entropy with 0.2 label smoothing.

We decompose the 100-million parallel documents corresponding to the 137 language pairs that include English and mine over 1B unique parallel sentences after filtering. After training models for each direction, we then evaluate the quality of the learned NMT models on a publicly available data set consisting of transcribed and translated TED talks~\cite{qi2018and}. Since the development and test sets were already tokenized, we first detokenize them using the Moses de-tokenizer. We performed additional checks to ensure that the TED test set isn't present in mined data.\footnote{Only a handful of high-frequency, short sentences present in the TED dataset were found in the mined data. Upon manual inspection at the source documents, we concluded that these do not constitute data leakage.}

In Table~\ref{tab:results_mt}, we report the BLEU scores from the mined bitexts (CCAligned) from our aligned documents on the TED talk \dataset{}. We include test set BLEU scores to a \dataset{} mined from Wikipedia (WikiMatrix)~\cite{schwenk2019wikimatrix} using LASER sentence embedding and margin-based sentence alignment as well as a cleaned paracrawl dataset (ParaCrawl v6) for comparison. Experimental conditions including model hyper-parameters between these NMT experiments were held constant making the BLEU scores directly comparable. As seen in the table, parallel sentences mined from CCAligned result in comparable BLEU scores to ParaCrawl v6 while yielding higher BLEU scores than WikiMatrix. 

While these results do not indicate superiority of one mining method over another (as there are significant differences in the number of mined sentences and nature of the corpus cleaning steps), they indicate that our mined document pairs are a valuable source of parallel data.

Moreover, the resulting corpus can be seen as a significant expansion to the coverage of Paracrawl. For instance, while the ParaCrawl consists of 23 European language pairs (paired with English), our aligned documents cover 8144 language pairs with 137 language pairs that include English. As such, this dataset has the potential to be mined for many low-resource language pairs not available in ParaCrawl. In addition, this dataset can be considered complementary to ParaCrawl as ParaCrawl incorporates techniques to align documents based on content which are then mined for parallel bitexts. Our aligned documents can be a valuable benchmark for ParaCrawl to leverage as they scale to additional, non-European language pairs.

\section{Conclusion \& Future Works}
In this paper, we apply URL-matching to curate a high-quality cross-lingual documents dataset from the CommonCrawl corpus. Our dataset contains over 392 million document pairs from 8144 language pairs covering 138 distinct languages. We first directly evaluate the quality of the URL-aligned pairs using human annotators.  We then introduce and evaluate simple embedding-based baseline techniques for aligning documents based on content. Our results indicate there is further work to be done to improve document alignment, especially for low-resource languages and that intelligent alignment schemes can significantly improve alignment performance across many language directions. Finally, we perform a case-study showing that our URL-aligned documents can be mined for high-quality parallel sentences which can be used to train machine translation models. Given the sheer size of this dataset, this has the potential to provide high-quality training data for many low-resource language pairs. 

One natural followup to this work is to develop techniques to better mine parallel sentences from these aligned documents -- especially for low-resource language pairs. Additional work could also leverage aligned documents as supervision to learn better cross-lingual document representations. Finally, while the aligned dataset is high-precision, leveraging this dataset for supervision in document alignment can potentially yield a larger, high-recall collection. To spur further work, we release the list of aligned URLs as well as code to generate aligned documents given CommonCrawl snapshots\footnote{Please contact Philipp Koehn for code or other resources. Any released data generated and shared can be found at \url{www.statmt.org/cc-aligned}}.

\balance
\bibliography{ccdata}
\bibliographystyle{acl_natbib}


\end{document}